\documentclass{article}

\usepackage{amsmath,amsfonts,bm}

\def\eqref#1{equation~\ref{#1}}

\def\1{\bm{1}}

\DeclareMathAlphabet{\mathsfit}{\encodingdefault}{\sfdefault}{m}{sl}
\SetMathAlphabet{\mathsfit}{bold}{\encodingdefault}{\sfdefault}{bx}{n}

\newcommand{\R}{\mathbb{R}}

\usepackage{hyperref}
\usepackage{url}
\usepackage[utf8]{inputenc} 
\usepackage[T1]{fontenc}    
\usepackage{hyperref}       
\usepackage{url}            
\usepackage{booktabs}       
\usepackage{amsfonts}       
\usepackage{nicefrac}       
\usepackage{microtype}      
\usepackage{xcolor}         
\usepackage{makecell}
\usepackage{graphicx}
\usepackage{amsmath}
\usepackage{multirow}
\usepackage{wrapfig}
\usepackage{wrapfig}
\usepackage{float}
\usepackage{graphicx}
\usepackage{caption}
\usepackage{color}
\usepackage{colortbl}

\def\ie{\emph{i.e., }}

\usepackage{microtype}
\usepackage{graphicx}
\usepackage{subfigure}
\usepackage{booktabs} 

\usepackage{hyperref}

\usepackage[accepted]{icml2024}

\usepackage{amsmath}
\usepackage{amssymb}
\usepackage{mathtools}
\usepackage{amsthm}

\usepackage[capitalize,noabbrev]{cleveref}

\theoremstyle{plain}

\theoremstyle{definition}

\theoremstyle{remark}

\usepackage[textsize=tiny]{todonotes}

\icmltitlerunning{Lightning Attention-2}

\begin{document}

\twocolumn[
\icmltitle{Lightning Attention-2: A Free Lunch for Handling Unlimited Sequence Lengths in Large Language Models}



\begin{icmlauthorlist}
\icmlauthor{Zhen Qin}{lab}
\icmlauthor{Weigao Sun}{lab}
\icmlauthor{Dong Li}{lab}
\icmlauthor{Xuyang Shen}{lab}
\icmlauthor{Weixuan Sun}{lab}
\icmlauthor{Yiran Zhong}{lab}
\end{icmlauthorlist}

\icmlaffiliation{lab}{OpenNLPLab}

\icmlcorrespondingauthor{Yiran Zhong}{zhongyiran@gmail.com}

\icmlkeywords{Linear attention, Lightning attention, unlimited sequence length, large language model}

\vskip 0.3in
]



\printAffiliationsAndNotice{}  

\begin{abstract}
Linear attention is an efficient attention mechanism that has recently emerged as a promising alternative to conventional softmax attention. With its ability to process tokens in linear computational complexities, linear attention, in theory, can handle sequences of unlimited length without sacrificing speed, \emph{i.e.,} maintaining a constant training speed for various sequence lengths with a fixed memory consumption. 
However, due to the issue with cumulative summation (\texttt{cumsum}), current Linear Attention algorithms cannot demonstrate their theoretical advantage in a casual setting. In this paper, we present Lightning Attention-2, the first linear attention implementation that enables linear attention to realize its theoretical computational benefits. To achieve this, we leverage the thought of tiling, separately handling the intra-block and inter-block components in linear attention calculation. Specifically, we utilize the conventional attention computation mechanism for the intra-blocks and apply linear attention kernel tricks for the inter-blocks. A tiling technique is adopted through both forward and backward procedures to take full advantage of the GPU hardware. We implement our algorithm in Triton to make it IO-aware and hardware-friendly. Various experiments are conducted on different model sizes and sequence lengths. Lightning Attention-2 retains consistent training and inference speed regardless of input sequence length and is significantly faster than other attention mechanisms. The source code is available at \href{https://github.com/OpenNLPLab/lightning-attention}{Lightning Attention-2}.
\end{abstract}

\section{Introduction}
\label{sec: intro}

\begin{figure*}[t]
    \centering
\includegraphics[width=1\textwidth]{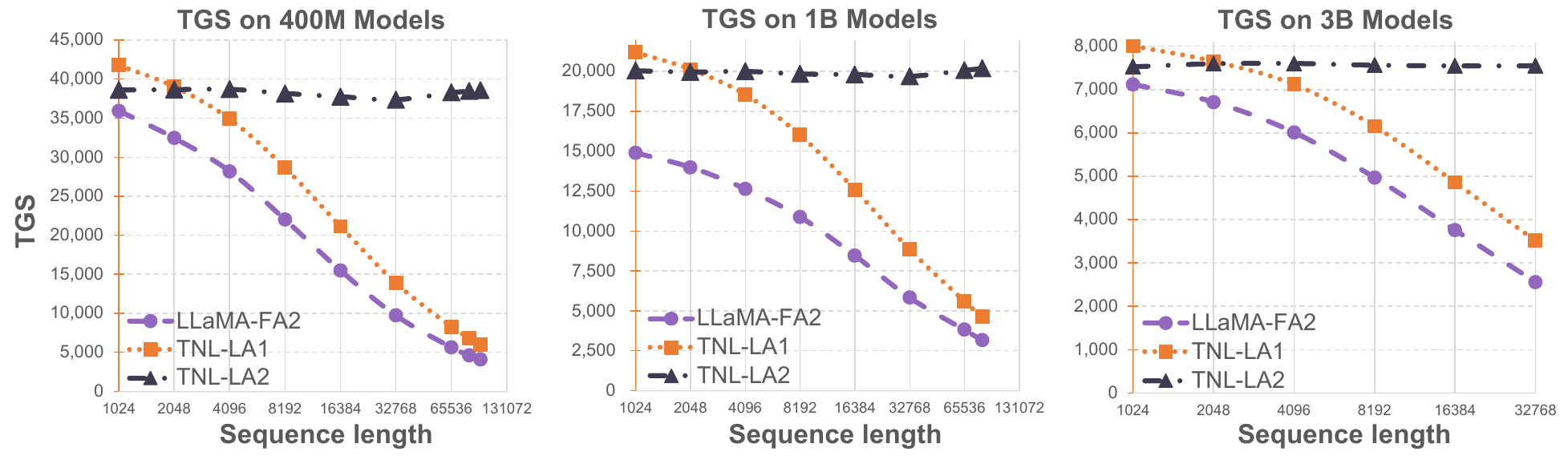}
    \vspace{-9mm}
    \caption{\textbf{Speed Showdown: FlashAttention vs. Lightning Attention in Expanding Sequence Lengths and Model Sizes.} The diagram above provides a comparative illustration of training speed, Token per GPU per Second (TGS) for LLaMA with FlashAttention-2, TransNormerLLM with Lightning Attention-1 and TransNormerLLM with Lightning Attention-2, implemented across three model sizes: 400M, 1B, and 3B from left to right. It is strikingly evident that Lightning Attention-2 manifests a consistent training speed irrespective of the increasing sequence length. Conversely, the other methods significantly decline training speed as the sequence length expands.}
    \label{fig: tgs}
\end{figure*}

The Transformer architecture has achieved widespread adoption, particularly in the domain of large language models (LLM)~\cite{brown2020language, touvron2023llama, 2307.09288, peng-etal-2023-rwkv,qin2023scaling} and multi-modal models~\cite{pmlr-v162-li22n, li2023blip2, liu2023visual, radford2021learning, li-etal-2023-map, lu2022linear, Mao_2023_ICCV, Shen_2023_CVPR,zhou2023audiovisual,10149455, hao2023improving}. However, its computational complexity grows quadratically with the length of the input sequence, making it challenging to model extremely long sequences.

Unlimited sequence length stands out as a noteworthy aspect within the realm of LLM, attracting considerable attention from researchers who seek intelligent solutions. The potential applications of LLM with unlimited sequence length are diverse, encompassing extended conversations in various professional domains and handling a vast number of tokens in multimodal modeling tasks.

In response to the quadratic complexity challenge, a promising resolution emerges in the form of linear attention. This method involves the elimination of the softmax operation and capitalizes on the associativity property of matrix products. Consequently, it significantly accelerates both training and inference procedures. To elaborate, linear attention reduces the computational complexity from $O(n^2)$ to $O(n)$ by leveraging the kernel trick~\cite{xfmrsarernns,Choromanski2020RethinkingAW,rfa,zhen2022cosformer} to compute the attention matrices, where $n$ represents the sequence length. This avenue holds substantial promise for augmenting the efficiency of transformer-style models across a broad spectrum of applications.

It is important to note that the notable reduction in complexity from $O(n^2)$ to $O(n)$ in linear attention is only theoretical and may not directly translate to a proportional improvement in computational efficiency on hardware in practice.
The realization of practical wall-clock speedup faces challenges, primarily stemming from two issues: 1). the dominance of memory access (I/O) on the GPU could impact the overall computation speed of attention. 2). the cumulative summation (\texttt{cumsum}) needed by the linear attention kernel trick prevents it from reaching its theoretical training speed in the causal setting. 

The first issue has been successfully addressed by Lightning Attention-1~\cite{qin2023scaling}. In this paper, we introduce \textbf{Lightning Attention-2} to solve the second issue. The key idea is to leverage the concept of "divide and conquer" by separately handling the intra block and inter block components in linear attention calculation. Specifically, for the intra blocks, we maintain the use of conventional attention computation mechanism to compute the product of $\mathbf{QKV}$, while for the inter blocks, we employ the linear attention kernel trick~\cite{xfmrsarernns}.
Tiling techniques are implemented in both forward and backward procedures to fully leverage GPU hardware capabilities. As a result, the Lightning Attention-2 can train LLMs with unlimited sequence length without extra cost\footnote{However, the sequence length may still be limited by hardware constraints, such as the GPU memory.}, as its computational speed remains constant with increasing sequence length under fixed memory consumption.

We performed a comprehensive evaluation of Lightning Attention-2 across a diverse range of sequence lengths to assess its accuracy and compare its computational speed and memory utilization with FlashAttention-2 ~\cite{dao2023flashattention2} and Lightning Attention-1. The findings indicate that Lightning Attention-2 exhibits a notable advantage in computational speed, attributed to its innovative intra-inter separation strategy. Additionally, Lightning Attention-2 demonstrates a reduced memory footprint compared to its counterparts without compromising performance.

\section{Related Work}
\label{sec: preliminary}
\subsection{Linear Attention}
Linear Transformer architectures discard the Softmax Attention mechanism, replacing it with distinct approximations~\citep{katharopoulos2020transformers,Choromanski2020RethinkingAW,rfa,zhen2022cosformer,qin-etal-2022-devil}. The key idea is to leverage the ``kernel trick" to accelerate the attention matrix computation, \ie~compute the product of keys and values first to circumvent the $n \times n$ matrix multiplication. Multiple methods have been proposed to replace the softmax operation. For instance, ~\citet{katharopoulos2020transformers} employ the $1+\mathrm{elu}$ activation function, ~\citet{zhen2022cosformer} utilize the cosine function to approximate softmax properties, and~\citet{ke2021rethinking,zheng2022linear,zheng2023efficient} leverage sampling strategies to directly mimic softmax operation. Despite having a theoretical complexity of $O(nd^2)$, the practical computational efficiency of linear attention diminishes notably in causal attention scenarios, primarily due to the necessity for \texttt{cumsum} operations~\citep{hua2022transformer}.

\subsection{IO-aware Attention}

The FlashAttention series~\citep{dao2022flashattention,dao2023flashattention2} focuses on system-level optimizations for the efficient implementation of the standard attention operator on GPU platforms. Extensive validation has demonstrated its effectiveness. The approach employs tiling strategies to minimize the volume of memory reads/writes between the GPU's high bandwidth memory (HBM) and on-chip SRAM.

To address the issue of slow computation for Linear Attention in the causal setting, Lightning Attention 1~\citep{qin2023scaling} employs the approach of FlashAttention-1/2, which involves segmenting the inputs $\mathbf{Q}, \mathbf{K}, \mathbf{V}$ into blocks, transferring them from slow HBM to fast SRAM, and then computing the attention output with respect to these blocks. Subsequently, the final results are accumulated. Although this method is much more efficient than the PyTorch implementation, it does not take advantage of the computational characteristics inherent to Linear Attention, and the theoretical complexity remains $O(n^2d)$.

\subsection{Long Sequence Handling in LLM}
A widely adopted strategy to tackle challenges related to length extrapolation involves the integration of Relative Positional Encoding (RPE) techniques \citep{su2021roformer,qin2023linearized}, strategically directing attention towards neighboring tokens. ALiBi~\citep{alibi} utilizes linear decay biases in attention mechanisms to mitigate the impact of distant tokens. Roformer~\citep{su2021roformer} introduces a novel Rotary Position Embedding (RoPE) method, widely embraced in the community, effectively leveraging positional information for transformer-based language model learning. Kerple~\citep{chi2022kerple} explores shift-invariant conditionally positive definite kernels within RPEs, introducing a suite of kernels aimed at enhancing length extrapolation properties, with ALiBi recognized as one of its instances. Furthermore, Sandwich~\citep{chi2023dissecting} postulates a hypothesis elucidating the mechanism behind ALiBi, empirically validating it by incorporating the hypothesis into sinusoidal positional embeddings. \citep{qin2023exploring} explored the sufficient conditions for additive relative position encoding to have extrapolation capabilities.

Instead of investigating the length extrapolation capability of transformers, some works also attempt to directly increase the context window sizes.
\citet{chen2023extending} introduces Position Interpolation (PI), extending context window sizes of RoPE-based pretrained Large Language Models (LLMs) such as LLaMA models to up to 32768 with minimal fine-tuning (within 1000 steps). StreamingLLM~\citep{xiao2023efficient} proposes leveraging the attention sink phenomenon, maintaining the Key and Value information of initial tokens to substantially recover the performance of window attention. As the sequence grows longer, the performance degrades. These methods can only extend sequence length in fine-tuning or testing phases, while our method allows training models in long sequence lengths from scratch with no additional cost.

\section{Method}
\label{sec: algo}
\subsection{Preliminary}
We first recall the formulation of linear attention and then introduce our proposed Lightning Attention-2.
In the case of NormAttention within TransNormer~\citep{qin-etal-2022-devil}, attention computation deviates from the conventional Transformer structure~\cite{vaswani2017attention} by eschewing the costly softmax and scaling operations. The NormAttention mechanism can be expressed as follows:
\begin{equation}
\mathbf{O}=\mathrm{Norm}((\mathbf{Q} \mathbf{K}^{\top})\mathbf{V}),
\label{eq: norm attention}
\end{equation}
where $\mathbf{Q}$, $\mathbf{K}$, and $\mathbf{V} \in \R^{n\times d}$ are the query, key, and value matrices, respectively, with $n$ denoting sequence length and $d$ representing feature dimension. To Leverage the computational efficiency inherent in right matrix multiplication, the above equation can be seamlessly and mathematically equivalently transformed into its linear variant, as dictated by the properties of matrix multiplication:
\begin{equation}
\mathbf{O}=\mathrm{Norm}(\mathbf{Q} (\mathbf{K}^{\top}\mathbf{V})),
\label{eq: norm attention 2}
\end{equation}
This linear formulation facilitates recurrent prediction with a commendable complexity of $O(nd^2),$ rendering it efficient during training relative to sequence length. Furthermore, employing linear attention ensures a constant computation complexity of $O(d^2)$ irrespective of sequence length, thereby enabling inference over unlimited long sequences. This achievement is realized by updating $\mathbf{K}^{\top}\mathbf{V}$ recurrently without the need for repeated computation of the entire attention matrix. In contrast, the standard softmax attention entails a computational complexity of $O(md^2)$ during the inference process, where $m$ denotes the token index.

Nevertheless, when dealing with causal prediction tasks, the effectiveness of the right product is compromised, leading to the requirement for the computation of \texttt{cumsum}~\citep{hua2022transformer}. This impediment hinders the potential for highly efficient parallel computation. Consequently, we persist with the conventional left matrix multiplication in Lightning Attention-1. This serves as the promotion behind the introduction of Lightning Attention-2, specifically crafted to address the challenges associated with the right product in such contexts.

\subsection{Lightning Attention-2}

\begin{figure}[t]
  \centering
  \includegraphics[width=1\columnwidth]{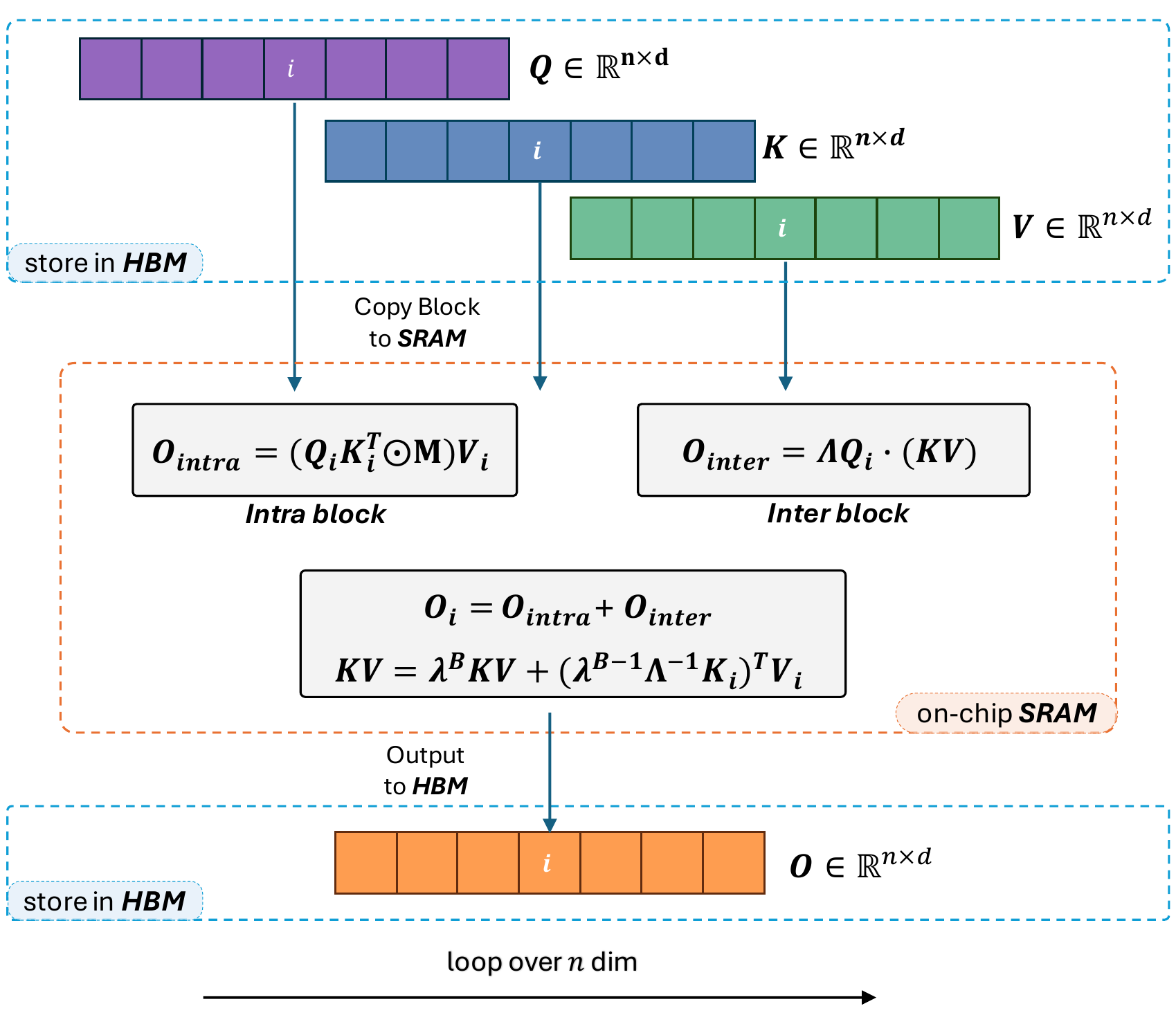}
  \vspace{-8mm}
  \caption{\textbf{Structural framework of Lightning Attention-2} is detailed in its algorithmic schematic. During the $i$-th iteration, the tiling blocks of matrices $\mathbf{Q}_i, \mathbf{K}_i, \mathbf{V}_i$ are transferred from High Bandwidth Memory (HBM) to Static Random-Access Memory (SRAM). Within the SRAM, the outputs $\mathbf{O}_{\mathrm{intra}}$ and $\mathbf{O}_{\mathrm{inter}}$ are computed independently, followed by an update to the $\mathbf{KV}$ matrix. Subsequently, the final output $\mathbf{O}_i$, which is the sum of $\mathbf{O}_{\mathrm{intra}}$ and $\mathbf{O}_{\mathrm{inter}}$, is written back from SRAM to HBM.}
  \label{fig:lightning2}
\end{figure}

Lightning Attention-2 employs a tiling methodology throughout its whole computation process. Given the huge variance in memory bandwidth between HBM and SRAM within GPU, Lightning Attention-2 applies a distinct strategy for 
leveraging them. In each iteration $i$, matrices $\mathbf{Q}_i, \mathbf{K}_i, \mathbf{V}_i$ undergo segmentation into blocks, subsequently transferred to SRAM for computation. The intra- and inter-block operations are segregated, with intra-blocks employing the left product and inter-blocks utilizing the right product. This approach optimally exploits the computational and memory efficiencies associated with the right product, enhancing overall execution speed. The intermediate activation $\mathbf{KV}$ is iteratively saved and accumulated within SRAM. Subsequently, the outputs of intra-blocks and inter-blocks are summed within SRAM, and the results are written back to HBM. This method aims to capitalize on the distinct advantages of each memory component, optimizing the computational workflow. The structural framework of Lightning Attention-2 is well illustrated in Fig.~\ref{fig:lightning2}.

The intricate details of the Lightning Attention-2 implementation are explicated through Algorithm~\ref{algo:Lightning Attention-2 fw pseudo} (forward pass) and Algorithm~\ref{algo:Lightning Attention-2 bw pseudo} (backward pass). These algorithms serve to encapsulate the nuanced computational procedures integral to Lightning Attention-2. Additionally, we provide a comprehensive derivation to facilitate a more profound comprehension of Lightning Attention-2. The derivations are systematically presented for both the forward pass and the backward pass, contributing to a thorough understanding of the underlying mechanisms.

\begin{algorithm}[t]
\small
    \caption{Lightning Attention-2 Forward Pass}
    \label{algo:Lightning Attention-2 fw pseudo}
    \begin{algorithmic}
    \STATE{\textbf{Input:} $\mathbf Q,\mathbf K,\mathbf V \in \mathbb{R}^{n \times d}$, decay rate $\lambda \in \mathbb R^+$, block sizes $B$.}
    \STATE{Divide $\mathbf {X}$ into $T = \frac{n}{B}$ blocks $\mathbf X_1, \mathbf X_2, ...\mathbf X_{T}$ of size $B \times d$ each, where $\mathbf X\in \{\mathbf Q, \mathbf K, \mathbf V,\mathbf O \}$. }
     \STATE{Initialize mask $\mathbf M\in \mathbb R^{B\times B}$, where $\mathbf M_{ij} = \lambda^{i-j}$, if $i\ge j$, else 0.}
     \STATE{Initialize $\Lambda =\mathrm{diag}\{\lambda, \lambda^2,\ldots, \lambda^{B}\}\in \mathbb R^{B\times B} $.} 
     \STATE{Initialize $\mathbf {KV} =0\in \mathbb R^{d\times d} $.}
    \FOR{$1 \leq i \leq T$}
        \STATE{Load $\mathbf Q_i,\mathbf  K_i, \mathbf V_i \in \mathbb{R}^{B \times d}$ from HBM to on-chip SRAM.}
        \STATE{On chip, compute $\mathbf O_{\mathrm{intra}}= [(\mathbf Q_i \mathbf K_i^{\top }) \odot \mathbf M]\mathbf V_i$.}
        \STATE{On chip, compute $\mathbf{O}_{\mathrm{inter}} =\Lambda \mathbf Q_i (\mathbf {KV}) $.}
        \STATE{On chip, compute $\mathbf{KV} =\lambda^B \mathbf{KV}+ (\lambda^{B}\Lambda^{-1} \mathbf K_i)^{\top}  \mathbf V_i$.}
      \STATE{Write $\mathbf O_i=\mathbf O_{\mathrm{intra}}+ \mathbf{O}_{\mathrm{inter}}$ to HBM as the $i$-th block of $\mathbf O$.}
      \ENDFOR
      \STATE{return $\mathbf O$.}
\end{algorithmic}
\end{algorithm}

\begin{algorithm}[t]
\small
    \caption{Lightning Attention-2 Backward Pass}
    \label{algo:Lightning Attention-2 bw pseudo}
    \begin{algorithmic}
    \STATE{\textbf{Input:} $\mathbf Q,\mathbf K,\mathbf V,\mathbf{dO} \in \mathbb{R}^{n \times d}$, decay rate $\lambda \in \mathbb R^+$, block sizes $B$.}
    \STATE{Divide $\mathbf {X}$ into $T = \frac{n}{B}$ blocks $\mathbf X_1, \mathbf X_2, ...\mathbf X_{T}$ of size $B \times d$ each, where $\mathbf X\in \{\mathbf Q, \mathbf K, \mathbf V \}$. }
     \STATE{Divide $\mathbf {dX}$ into $T = \frac{n}{B}$ blocks $\mathbf {dX}_1, \mathbf {dX}_2, ...\mathbf {dX}_{T}$ of size $B \times d$ each, where $\mathbf X\in \{\mathbf Q, \mathbf K, \mathbf V, \mathbf O  \}$ }.
     \STATE{Initialize mask $\mathbf M\in \mathbb R^{B\times B}$, where $\mathbf M_{ij} = \lambda^{i-j}$, if $i\ge j$, else 0.}
     \STATE{Initialize $\Lambda =\mathrm{diag}\{\lambda, \lambda^2,\ldots, \lambda^{B}\}\in \mathbb R^{B\times B} $} .
      \STATE{Initialize $\mathbf {KV} =0, \mathbf{dKV}=0\in \mathbb R^{d\times d} $.}

  \FOR{$i=  1,\ldots ,T$}
        \STATE{Load $\mathbf K_i, \mathbf V_i, \mathbf O_i, \mathbf {dO}_i \in \mathbb{R}^{B \times d}$ from HBM to on-chip SRAM.}
    
        \STATE{On chip, compute $\mathbf {dQ}_{\mathrm{intra}} =[(\mathbf {dO}_i \mathbf V_i^{\top}) \odot \mathbf M] \mathbf{K}_i$.}

    \STATE{On chip, compute $\mathbf {dQ}_{\mathrm{inter}} =\Lambda \mathbf{dO}_i (\mathbf{KV})^{\top} $.}

     \STATE{On chip, compute $\mathbf{KV} =\lambda^B \mathbf{KV}+ (\lambda^{B}\Lambda^{-1} \mathbf K_i)^{\top}  \mathbf V_i$.}

     \STATE{Write $\mathbf {dQ}_i=\mathbf{dQ}_{\mathrm{intra}} + \mathbf{dQ}_{\mathrm{inter}}$ to HBM as the $i$-th block of $\mathbf {dQ}$.}
      \ENDFOR

    \FOR{$i=  T,\ldots ,1$}
        \STATE{Load $\mathbf Q_i, \mathbf K_i, \mathbf V_i, \mathbf O_i, \mathbf {dO}_i \in \mathbb{R}^{B \times d}$ from HBM to on-chip SRAM.}
        
    \STATE{On chip, compute $\mathbf{dK_{\mathrm{intra}}}=[(\mathbf {dO}_i \mathbf V_i^{\top}) \odot \mathbf M ]^{\top} \mathbf{Q}_i$.}
     \STATE{On chip, compute $\mathbf{dK_{\mathrm{inter}}}={(\lambda^{B} \Lambda^{-1} \mathbf V_i)}(\mathbf{dKV})^{\top}$.}
     
  \STATE{On chip, compute $\mathbf{dV_{\mathrm{intra}}}=[(\mathbf Q_i \mathbf K_i^{\top}) \odot \mathbf M ]^{\top} \mathbf{dO}_i$.}
     \STATE{On chip, compute $\mathbf{dV_{\mathrm{inter}}}=(\lambda^{B} \Lambda^{-1}\mathbf K_i) \mathbf{dKV}$.}
     
   \STATE{On chip, compute $\mathbf{dKV} =\lambda^B \mathbf{dKV}+ (\Lambda \mathbf Q_i)^{\top}  \mathbf {dO}_i $.}

     \STATE{Write $ \mathbf {dK}_i=\mathbf K_{\mathrm{intra}} +\mathbf K_{\mathrm{inter}} ,\mathbf {dV}_i=\mathbf V_{\mathrm{intra}} +\mathbf V_{\mathrm{inter}}$ to HBM as the $i$-th block of $ \mathbf {dK}, \mathbf {dV}$.}
      \ENDFOR
      \STATE{return $\mathbf {dQ, dK, dV}$.}
\end{algorithmic}
\end{algorithm}

\subsubsection{Forward Pass}
We ignore the $\text{Norm}(\cdot)$ operator in eq. (\ref{eq: norm attention 2}) to simplify the derivations.
During forward pass of Lightning Attention-2, the $t$-th output can be formulated as
\begin{equation}
\mathbf{o}_t = \mathbf{q}_t \sum_{s \le t} \lambda^{t-s} \mathbf{k}_s^{\top} \mathbf{v}_s.
\end{equation}

In a recursive form, the above equation can be rewritten as
\begin{equation}
\begin{aligned}
\mathbf{kv}_0 &= 0 \in \mathbb{R}^{d \times d}, \\
\mathbf{kv}_t &= \lambda \mathbf{kv}_{t-1} + \mathbf{k}_t^{\top} \mathbf{v}_t, \\
\mathbf{o}_t &= \mathbf{q}_t (\mathbf{kv}_t),
\end{aligned}
\end{equation}
where
\begin{equation}
\mathbf{kv}_t = \sum_{s \le t} \lambda^{t-s} \mathbf{k}_s^{\top} \mathbf{v}_s.
\end{equation}
To perform tiling, let us write the equations in block form. Given the total sequence length $n$ and block size $B$, $\mathbf{X}$ is divided into $T = \frac{n}{B}$ blocks $\{\mathbf{X}_1, \mathbf{X}_2, \ldots, \mathbf{X}_T\}$ of size $B \times d$ each, where $\mathbf{X} \in \{\mathbf{Q}, \mathbf{K}, \mathbf{V}, \mathbf{O}\}$. 

We first define
\begin{equation}
\mathbf{KV}_0 =\mathbf 0\in \mathbb{R}^{d \times d}, \\
\mathbf{KV}_t = \sum_{s \le tB} \lambda^{tB-s} \mathbf{k}_s^{\top} \mathbf{v}_s.
\end{equation}
Given $\mathbf{KV}_t$, the output of $(t+1)$-th block, i.e., $tB+r$, with $1 \le r \le B$ is
\begin{equation}
\begin{aligned}
&\mathbf{o}_{tB+r} \\
= &\mathbf{q}_{tB+r} \sum_{s \le tB+r} \lambda^{tB+r-s} \mathbf{k}_s^{\top} \mathbf{v}_s \\
= &\mathbf{q}_{tB+r}\left( \sum_{s=tB+1}^{tB+r} \lambda^{tB+r-s} \mathbf{k}_s^{\top} \mathbf{v}_s + \lambda^r \sum_{s \le tB} \lambda^{tB-s} \mathbf{k}_s^{\top} \mathbf{v}_s \right) \\
= & \mathbf{q}_{tB+r} \sum_{s=tB+1}^{tB+r} \lambda^{tB+r-s} \mathbf{k}_s^{\top} \mathbf{v}_s + \lambda^r \mathbf{q}_{tB+r} \mathbf{kv}_{tB} .
\end{aligned}
\end{equation}
Rewritten in matrix form, we have
\begin{equation}
\begin{aligned}
\mathbf{O}_{t+1}=  &
\underbrace{[(\mathbf{Q}_{t+1} \mathbf{K}_{t+1}^{\top}) \odot \mathbf{M}] \mathbf{V}_{t+1}}_{\mathrm{Intra\ Block}} \\
&+ \underbrace{\Lambda\mathbf{Q}_{t+1} (\mathbf{KV}_t)}_{\mathrm{Inter\ Block}},\\
\end{aligned}
\end{equation}
where
\begin{equation}
\begin{aligned}
\mathbf{M}_{st} &= \begin{cases}
\lambda^{s-t} & s \ge t\\
0 & s < t
\end{cases} , \\
\Lambda&=\mathrm{diag}\{1, \ldots, \lambda^{B-1}\} .
\end{aligned}
\end{equation}
And the $\mathbf{KV}$ at $(t+1)$-th block can be written as
\begin{equation}
\begin{aligned}
\mathbf{KV}_{t+1} &= \sum_{s \le (t+1)B} \lambda^{(t+1)B-s} \mathbf{k}_s^{\top} \mathbf{v}_s \\
&= \lambda^B \sum_{s \le tB} \lambda^{tB-s} \mathbf{k}_s^{\top} \mathbf{v}_s + \sum_{s=tB+1}^{(t+1)B} \lambda^{(t+1)B-s} \mathbf{k}_s^{\top} \mathbf{v}_s \\
&= \lambda^B \mathbf{KV}_t + \left(\mathrm{diag}\{\lambda^{B-1}, \ldots, 1\} \mathbf{K}_{t}\right)^{\top} \mathbf{V}_{t} \\
&= \lambda^B \mathbf{KV}_t + \left(\lambda^{B}\Lambda^{-1} \mathbf{K}_{t}\right)^{\top} \mathbf{V}_{t}.
\end{aligned}
\end{equation}
The complete expression of the forward pass of Lightning Attention-2 can be found in Algorithm~\ref{algo:Lightning Attention-2 fw pseudo}.

\subsubsection{Backward Pass}
For backward pass, let us consider the reverse process.
First given $\mathbf{do}_t$, we have
\begin{equation}
\begin{aligned}
\mathbf{dq}_t &= \mathbf{do}_t (\mathbf{kv}_t)^\top \in \mathbb{R}^{1 \times d}, \\
\mathbf{dk}_t &= \mathbf{v}_t (\mathbf{dkv}_t)^\top \in \mathbb{R}^{1 \times d}, \\
\mathbf{dv}_t &= \mathbf{k}_t (\mathbf{dkv}_t) \in \mathbb{R}^{1 \times d}, \\
\mathbf{dkv}_t &= \sum_{s \geq t} \lambda^{s-t} \mathbf{q}_s^\top \mathbf{do}_s \in \mathbb{R}^{d \times d}.
\end{aligned}
\end{equation}
By writing $\mathbf{dkv}_t$ in a recursive form, we get
\begin{equation}
\begin{aligned}
\mathbf{dkv}_{n+1} &= 0 \in \mathbb{R}^{d \times d}, \\\quad \mathbf{dkv}_{t-1} &= \lambda \mathbf{dkv}_t + \mathbf{q}_{t-1}^\top \mathbf{do}_{t-1}.
\end{aligned}
\end{equation}
To facilitate the understanding of tiling, let us consider the above equations in block style. Given the total sequence length $n$ and block size $B$, $\mathbf{X}$ is divided into $T = \frac{n}{B}$ blocks $\{\mathbf{X}_1, \mathbf{X}_2, \ldots, \mathbf{X}_T\}$ of size $B \times d$ each, where $\mathbf{X} \in \{ \mathbf{Q}, \mathbf{K}, \mathbf{V}, \mathbf{O}, \mathbf{dO} \}$. 

We first define
\begin{equation}
    \begin{aligned} 
    \mathbf{dKV}_{T+1}&=\mathbf 0 \in \mathbb{R}^{d \times d}, \\
    \mathbf{dKV}_t &= \sum_{s > tB} \lambda^{s-tB} \mathbf{q}_s^\top \mathbf{do}_s.
    \end{aligned}
\end{equation}

Then for the $(t+1)$-th block, i.e., $tB+r, 0 \leq r < B$, we have
\begin{equation}
\begin{aligned}
&\mathbf{dq}_{tB+r} \\
=& \mathbf{do}_{tB+r} \sum_{s \leq tB+r} \lambda^{tB+r-s} \mathbf{v}_s^\top \mathbf{k}_s \\
=& \mathbf{do}_{tB+r} \left(\sum_{s=tB+1}^{tB+r} \lambda^{tB+r-s} \mathbf{v}_s^\top \mathbf{k}_s 
+ \lambda^r \sum_{s \leq tB} \lambda^{tB-s} \mathbf{v}_s^\top \mathbf{k}_s 
\right) \\
=&\mathbf{do}_{tB+r} \sum_{s=tB+1}^{tB+r} \lambda^{tB+r-s} \mathbf{v}_s^\top \mathbf{k}_s +  \lambda^r \mathbf{do}_{tB+r} \mathbf{kv}_{tB}^\top .
\end{aligned}
\end{equation}
In matrix form, we have
\begin{equation}
\begin{aligned}
\mathbf{dQ}_{t+1} =&
\underbrace{[(\mathbf{dO}_{t+1} \mathbf{V}_{t+1}^\top) \odot \mathbf{M}] \mathbf{K}_{t+1}}_{{\mathrm{Intra\ Block}}} \\
&+ \underbrace{\Lambda \mathbf{dO}_{t+1} (\mathbf{KV}_t^\top)}_{{\mathrm{Inter\ Block}}}.
\end{aligned}
\end{equation}
Since the recursion of $\mathbf{dK}_t$ steps from $t+1$ to $t$, given $\mathbf{KV}_{t+1}$, $\mathbf{dK}_{t}$ for the $t$-th block, i.e., at positions $(t-1)B+r, 0< r \le B$ is
\begin{equation}
\begin{aligned}
&\mathbf{dk}_{(t-1)B+r} \\
= &\mathbf{v}_{(t-1)B+r} \sum_{s \geq (t-1)B+r} \lambda^{s-(t-1)B-r} \mathbf{do}_s^\top \mathbf{q}_s \\
=& \mathbf{v}_{(t-1)B+r} \left(
\sum_{s=(t-1)B+r}^{tB} \lambda^{tB+r-s} \mathbf{do}_s^\top \mathbf{q}_s \right) \\
&+
 \mathbf{v}_{(t-1)B+r}\left(\lambda^{B-r} \sum_{s > tB} \lambda^{s-tB} \mathbf{do}_s^\top \mathbf{q}_s \right)
\\
= &\mathbf{v}_{(t-1)B+r} \sum_{s=(t-1)B+r}^{tB} \lambda^{tB+r-s} \mathbf{do}_s^\top \mathbf{q}_s \\
&+ \lambda^{B-r} \mathbf{v}_{(t-1)B+r} \mathbf{dKV}_{t}^\top.
\end{aligned}
\end{equation}
In matrix form, we get
\begin{equation}
\begin{aligned}
\mathbf{dK}_{t-1} =&
\underbrace{[(\mathbf{dO}_{t-1} \mathbf{V}_{t-1}^\top) \odot \mathbf{M}]^\top \mathbf{Q}_{t-1}}_{{\mathrm{Intra\ Block}}} \\
&+ \underbrace{
\lambda^{B} \Lambda^{-1}
\mathbf{V}_{t-1} (\mathbf{dKV}_t^\top)}_{{\mathrm{Inter\ Block}}}.
\end{aligned}
\end{equation}

\begin{figure*}[htb]
\centering
\includegraphics[width=0.95\textwidth]{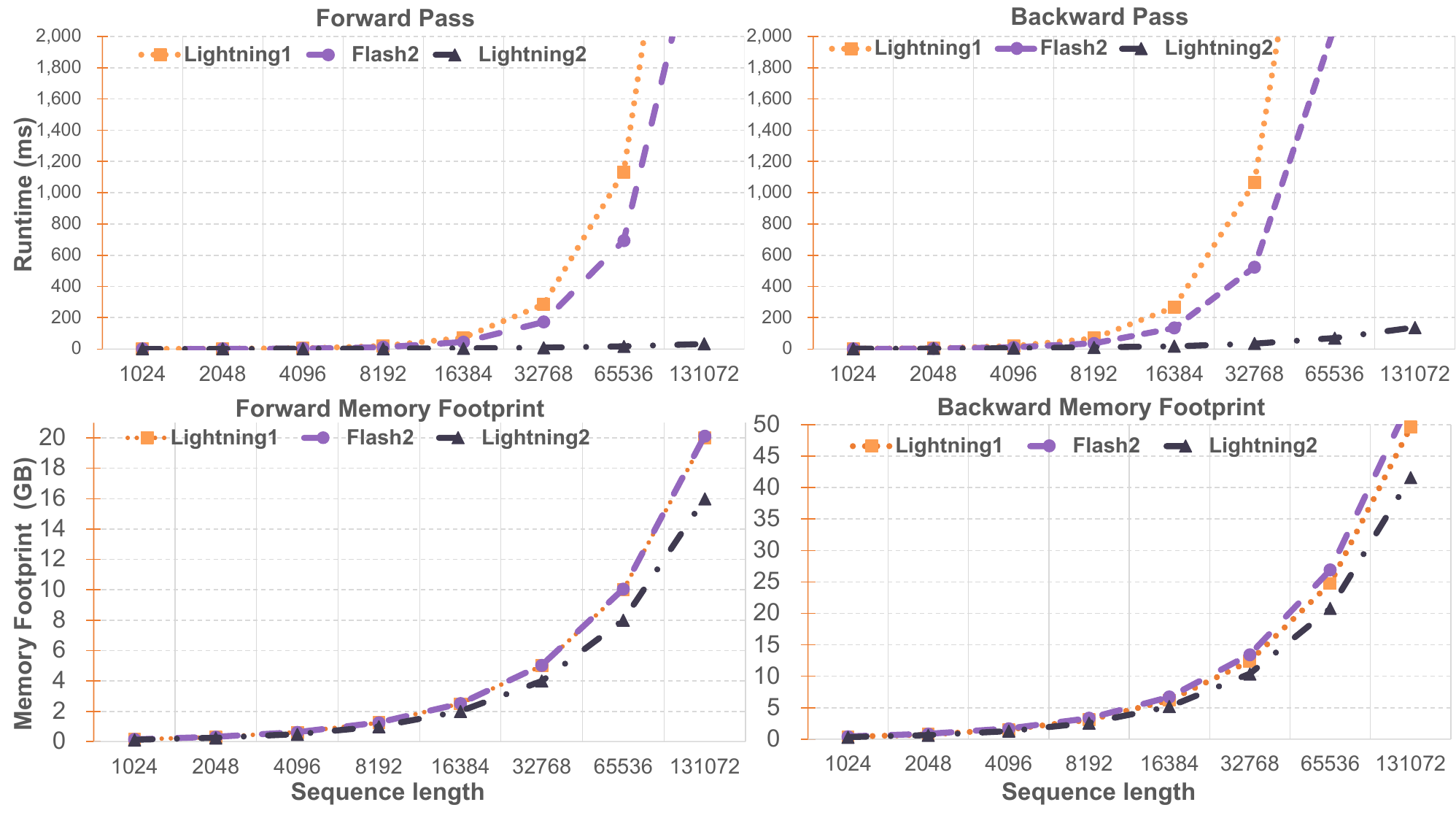}
\vspace{-4mm}
\caption{\textbf{Comparative Analysis of Speed and Memory Usage: FlashAttention vs. Lightning Attention.} Upper Section: Runtime in milliseconds for the forward and backward pass across varying sequence lengths. Lower Section: Memory utilization during the forward and backward pass at different sequence lengths.}
\vspace{-3mm}
\label{fig: timecomp}
\end{figure*}

Considering $\mathbf{dV}_t$ for the $t$-th block, i.e., at positions $(t-1)B+r, 0 < r \le B$, we have
\begin{equation}
\begin{aligned}
&\mathbf{dv}_{(t-1)B+r} \\
=& \mathbf{k}_{(t-1)B+r} \sum_{s \geq (t-1)B+r} \lambda^{s-(t-1)B-r} \mathbf{q}_s^\top \mathbf{do}_s \\
=& \mathbf{k}_{(t-1)B+r} \left(
\sum_{s=(t-1)B+r}^{tB} \lambda^{tB+r-s} \mathbf{q}_s^\top \mathbf{do}_s \right) \\
&+ \lambda^{B-r} \left(\sum_{s> tB} \lambda^{s-tB} \mathbf{q}_s^\top \mathbf{do}_s \right) \\
=& \mathbf{k}_{(t-1)B+r} \sum_{s=(t-1)B+r}^{tB} \lambda^{tB+r-s} \mathbf{q}_s^\top \mathbf{do}_s \\
&+ \lambda^{B-r} \mathbf{k}_{(t-1)B+r} \mathbf{dKV}_{t} .
\end{aligned}
\end{equation}
In matrix form, we get
\begin{equation}
\begin{aligned}
\mathbf{dV}_{t-1} =&
\underbrace{[(\mathbf{Q}_{t-1} \mathbf{K}_{t-1}^\top) \odot \mathbf{M}]^\top \mathbf{dO}_{t}}_{{\mathrm{Intra\ Block}}} \\
&+ \underbrace{\lambda^{B} \Lambda^{-1} \mathbf{K}_{t-1} (\mathbf{dKV}_t)}_{{\mathrm{Inter\ Block}}}.
\end{aligned}
\end{equation}
Finally, the recursive relation for $\mathbf{dKV}_t$ is 
\begin{equation}
\begin{aligned}
\mathbf{dKV}_{t}
&= \sum_{s > tB} \lambda^{s-tB} \mathbf{q}_s^\top \mathbf{do}_s \\
&= \lambda^B \sum_{s > (t+1)B} \lambda^{s-(t+1)B} \mathbf{q}_s^\top \mathbf{do}_s \\
&+ \sum_{s=tB+1}^{(t+1)B} \lambda^{s-tB} \mathbf{q}_s^\top \mathbf{do}_s \\
&= \lambda^B \mathbf{dKV}_{t+1} + \left(\Lambda \mathbf{Q}_t \right)^\top \mathbf{dO}_t.
\end{aligned}
\end{equation}
Algorithm~\ref{algo:Lightning Attention-2 bw pseudo} describes the backward pass of Lightning Attention-2 in more detail.

\paragraph{Discussion}
A recent method, GLA~\citep{2312.06635} models sequences using linear attention with data-dependent decay. Its chunk-wise Block-Parallel Algorithm employs tiling and IO-aware concepts. However, unlike Lightning Attention-2, it uses parallel computations for each block, which leads to higher memory usage.
Retnet~\citep{2307.08621} is very similar in structure to TransNormerLLM~\cite{qin2023scaling} and uses the chunk-wise retention algorithm. This algorithm is comparable to the forward pass of Lightning Attention-2 but does not consider IO-aware or the backward pass.

\section{Experiments}
\label{sec: exper}

\begin{figure*}[t]
    \centering
\includegraphics[width=1\textwidth]{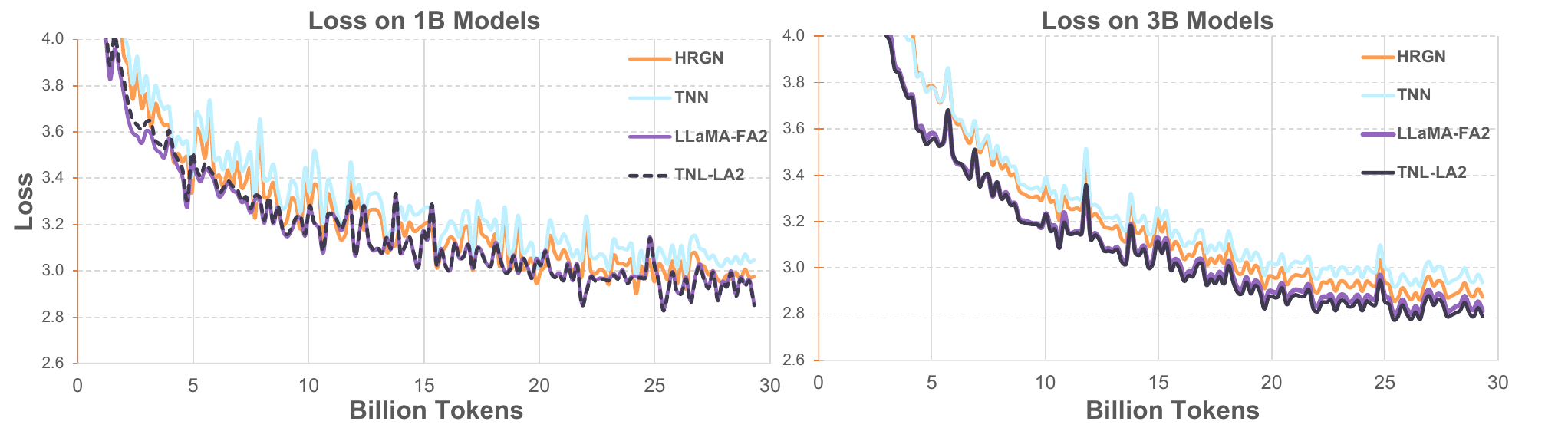}
    \vspace{-8mm}
    \caption{\textbf{Performance Comparison of HGRN, TNN, LLaMA with FlashAttention2 and TransNormerLLM with Lightning Attention-2.} For the 1B model, we used 16$\times$A800 80G GPUs with a batch size of 12 per GPU; for the 3B model, we scaled up to 32$\times$A800 80G GPUs and a batch size of 30 per GPU. The training context length was set to 2K.}
    \label{fig: loss}
     \vspace{-4mm}
\end{figure*}

\definecolor{Gray}{gray}{0.95}
\begin{table*}[!]
    \centering
   
    \small    
    \caption{\textbf{Efficiency Comparison of LLaMA with FlashAttention2, TransNormerLLM with Lightning Attention-1, and TransNormerLLM with Lightning Attention-2.} The statistical analysis was performed using 2$\times$A100 80G GPUs. The table reports Tokens per GPU per Second (TGS) across three different model sizes, within context ranges spanning from 1K to 92K. OOM stands for out of GPU memory.}
    \setlength{\tabcolsep}{3.3mm}
        \begin{tabular}{p{2cm}|c|ccccccccc}
        \toprule[0.8pt]
            Model & PS & 1024 & 2048 & 4096 & 8192 & 16384 & 32768 & 65536 & 81920 & 94208 \\  \midrule
                    LLaMA-FA2       & 0.4B & 35931 & 32453 & 28184 & 21996 & 15479 & 9715 & 5643 & 4604 & 4078 \\
                    TNL-LA1 & 0.4B & 41789 & 39043 & 34894 & 28627 & 21112 & 13852 & 8247 & 6824 & 6012 \\ 
    \rowcolor{Gray} TNL-LA2 & 0.4B & 38615 & 38680 & 38714 & 38172 & 37755 & 37364 & 38278 & 38457 & 38596\\ \hline
            LLaMA-FA2       & 1B   & 14897 & 13990 & 12644 & 10887 & 8468 & 5836 & 3820 & 3167 & OOM \\
            TNL-LA1 & 1B  & 21195 & 20128 & 18553 & 16012 & 12594 & 8848 & 5611 & 4625 & OOM \\ 
    \rowcolor{Gray} TNL-LA2 & 1B & 20052 & 19967 & 20009 & 19841 & 19805 & 19691 & 20077 & 20186 & OOM  \\ \hline
            LLaMA-FA2       & 3B & 7117 & 6708 & 6008 & 4968 & 3755 & 2558 & OOM & OOM & OOM   \\
            TNL-LA1 & 3B & 8001 & 7649 & 7117 & 6152 & 4859 & 3512 & OOM & OOM & OOM  \\ 
    \rowcolor{Gray} TNL-LA2 & 3B & 7524 & 7593 & 7599 & 7559 & 7545 & 7545  & OOM & OOM & OOM   \\

        \toprule[0.8pt]              
        \end{tabular}
 \label{tab:speed}
\vspace{-4mm}
\end{table*}

To comprehensively assess Lightning Attention-2's performance, speed, and memory utilization, we conducted extensive experiments on the TransNormerLLM model, with Lightning Attention-2 integrated. Our implementation utilizes the Metaseq framework~\citep{zhang2022opt}, a PyTorch-based sequence modeling framework~\citep{paszke2019pytorch}. All experiments are executed on the GPU cluster featuring 128 A100 80G GPUs. The deployment of Lightning Attention-2 is implemented in Triton~\citep{Tillet2019TritonAI}.

\subsection{Attention Module Evaluation}

We conducted a comparison of speed and memory usage among attention modules Lightning Attention-1, Lightning Attention-2, and FlashAttention-2, all under a single A100 80G GPU. As depicted in Figure~\ref{fig: timecomp}, the analysis focuses on the runtime, measured in milliseconds, for the separated forward and backward propagation. The baseline runtime demonstrates a quadratic growth relative to the sequence length. In contrast, Lightning Attention-2 exhibits a markedly superior performance with linear growth. Notably, as the sequence length increases, this disparity in runtime becomes increasingly apparent. In addition to speed enhancements, our method also maintains a significant advantage in memory usage with the increase in sequence length.

\subsection{Lightning Attention-2 in Large Language Model}

\begin{table}[H]
    \centering
    \small
    \setlength{\tabcolsep}{2mm}
    \vspace{-4mm}
    \caption{\textbf{Language Modeling Comparison} between TransNormerLLM with Lightning Attention-1 and Lightning Attention-2.} 
    \begin{tabular}{ccccc}
    \toprule[0.8pt]
        Model  & Attention & Params & Updates & Loss \\ \hline
        TNL-LA1 & {LA1} & 0.4B & 100k   & 2.229\\
        TNL-LA2 & {LA2} & 0.4B & 100k   & 2.228 \\
    \toprule[0.8pt]
    \end{tabular}
    \label{tab:performance}
     \vspace{-4mm}
\end{table}

\definecolor{Gray}{gray}{0.95}
\begin{table*}[t]
    \centering
    \small    
    \caption{\textbf{Performance Comparison on Commonsense Reasoning and Aggregated Benchmarks.} TNL-LA2: TransNormerLLM with Lightning Attention-2. PS: parameter size (billion). T: tokens (billion).
    HS: HellaSwag. WG: WinoGrande. }
    \setlength{\tabcolsep}{0.75mm}
        \begin{tabular}{p{1.32cm}|cc|cccccccc|ccccc}
        \toprule[0.8pt]
            Model & PS & T & BoolQ & PIQA & HS & WG & ARC-e & ARC-c & OBQA &CSR & C-Eval & MMLU & C-Eval & MMLU \\ \hline
             & B & B &acc &acc &acc\_norm &acc &acc &acc\_norm &acc\_norm &avg. &acc-0shot &acc-0shot &acc-5shot &acc-5shot \\ \midrule
             
                     Pythia & 12 & 50.3  &\textbf{62.14} &71.76 &51.89 &55.64 &59.22 &28.75 &\textbf{32.80} &51.74  &22.36 &25.80 &21.43 &26.10 \\
    \rowcolor{Gray} TNL-LA2 & 15 & 49.8  &62.08 &\textbf{72.52} &\textbf{55.55} &\textbf{57.14} &\textbf{62.12} &\textbf{31.14} &32.40 &\textbf{53.28}  &\textbf{25.55} &\textbf{26.60} &\textbf{26.18} &\textbf{27.50}  \\ \hline
                       Pythia & 12 & 100.6 &62.20 &73.23 &58.83 &59.35 &63.76 &31.91 &32.80 &54.58  &24.00 &24.80 &24.45 &24.40 \\
     \rowcolor{Gray}  TNL-LA2 & 15 & 99.7  &\textbf{63.98} &\textbf{74.70} &\textbf{61.09} &\textbf{61.33} &\textbf{65.95} &\textbf{34.64} &\textbf{35.60} &\textbf{56.76}  &\textbf{26.70} &\textbf{26.90} &\textbf{25.38} &\textbf{27.40}  \\
        \toprule[0.8pt]
              
        \end{tabular}
    \label{tab:benchmark}
\vspace{-3mm}
\end{table*}

\paragraph{Performance Evaluation} In Table~\ref{tab:performance}, we evaluated the performance of the TransNormerLLM-0.4B model under 2K contexts, comparing two variants: one equipped with Lightning Attention-1 and the other with Lightning Attention-2. These experiments were carried out using 8$\times$A100 80G GPUs. After 100,000 iterations, using the sampled corpus from our corpus with 300B tokens and initial seed, we observed a marginal performance difference. Specifically, the variant with Lightning Attention-2 demonstrated a performance decrement of 0.001 compared to its counterpart with Lightning Attention-1.

Furthermore, our analysis extended to benchmarking the top-tier efficient large language models, including LLaMA-FA2~\cite{touvron2023llama, dao2023flashattention2}, TNL-LA2, HGRN~\cite{qin2023hierarchically}, and TNN~\cite{qin2023toeplitz}. This benchmarking focused on training loss using a 30B subset of our uniquely assembled corpus, scaling from 1 to 3 billion parameters. As depicted in Figure~\ref{fig: loss}, the TNL-LA2 model achieved marginally lower loss compared to the other models under review in both 1B and 3B parameters.

\vspace{-2mm}
\paragraph{Efficiency Evaluation} In Table~\ref{tab:speed}, we present a comparative analysis of training speeds under the same corpora and hardware setups. This comparison encompasses three variants: TransNormerLLM with Lightning Attention-2 (TNL-LA2), TransNormerLLM with Lightning Attention-1 (TNL-LA1), and LLaMA with FlashAttention2 (LLaMA-FA2). Our findings show that during both the forward and backward passes, the TGS (tokens per GPU per second) for TNL-LA2 remains consistently high, while the other two models exhibit a rapid decline when the sequence length is scaled from 1K to 92K. This pattern suggests that Lightning Attention-2 offers a significant advancement in managing unlimited sequence lengths in LLM.

\subsection{Benchmarking Lightning Attention-2 in Large Language Model}
To evaluate the performance of the Lightning Attention-2, we conducted an analysis of the TransNormerLLM-15B~\cite{qin2023scaling}, a model comprising 15 billion parameters. The TransNormerLLM-15B is characterized by its 42 layers, 40 attention heads, and an overall embedding dimension of 5120. The model will be trained on a corpus of more than 1.3 trillion tokens with a sequence length of 6,144. Notably, the model achieved a processing speed of 1,620 tokens per GPU per second. Given that the comprehensive pre-training phase is scheduled to span three months, we hereby present the most recent results from the latest checkpoint for inclusion in Table~\ref{tab:benchmark}.

This evaluation is conducted using the lm-evaluation-harness framework~\cite{eval-harness}. Our benchmark focuses on two key areas: Commonsense Reasoning (CSR) and Multiple Choice Questions (MCQ). For comparative analysis, we also evaluated the Pythia-12B~\cite{biderman2023pythia} model under the same benchmarks.

\vspace{-2mm}
\paragraph{Commonsense Reasoning} We report BoolQ \citep{clark2019boolq}, PIQA \citep{bisk2019piqa}, SIQA \citep{sap2019socialiqa},
HellaSwag \citep{zellers2019hellaswag}, WinoGrande \citep{sakaguchi2019winogrande}, ARC easy and challenge \citep{clark2018think}, OpenBookQA \citep{mihaylov2018suit} and their average. In all CSR tasks, the performance of TransNormerLLM-15B surpassed Pythia-12B by about 2\%. Furthermore, TransNormerLLM-15B-100B showed an approximate 3.5\% improvement over its 50 billion-token stage, especially in the HellaSwag task, with over a 5\% performance increase.

\vspace{-2mm}
\paragraph{Aggregated Benchmarks}  We report the overall results for MMLU \citep{hendrycks2021measuring} and C-Eval \citep{huang2023ceval} with both 0-shot and 5-shot settings. In the C-Eval tasks, TransNormerLLM-15B is about 2\% higher than Pythia-12B. In the 0-shot and 5-shot tests in both Chinese (C-Eval) and English (MMLU), TransNormerLLM-15B's performance also exceeded the 25\% baseline (the probability of random selection in a 4-choice scenario). We also noticed fluctuations in the 5-shot MCQ tasks, with an average MCQ score of around 26.5\%.

\section{Conclusion}
\label{sec: conclusion}
In this paper, we introduced Lightning Attention-2, a pioneering implementation of linear attention that effectively harnesses its theoretical computational advantages, particularly in the causal setting. Our approach, which adopts the concepts of "divide and conquer" and tiling techniques, successfully addresses the limitations of current linear attention algorithms, especially the challenges associated with cumulative summation. By separating the computation into intra-block and inter-block components, we effectively leverage GPU hardware to its fullest potential, ensuring efficiency. Our extensive experiments across various model sizes and sequence lengths demonstrate that Lightning Attention-2 not only maintains consistent training speeds regardless of input sequence length but also outperforms existing state-of-the-art attention mechanisms in terms of speed and accuracy. This breakthrough has profound implications for the future of large language models, particularly those requiring the processing of long sequences. Looking ahead, we intend to introduce sequence parallelism in conjunction with Lightning Attention-2, which aims to facilitate the training of extra-long sequences, effectively overcoming existing hardware constraints.

\section*{Acknowledgement}
This work is partially supported by the National Key R\&D Program of China (NO.2022ZD0160100). We thank Songlin Yang for the helpful discussions.

\bibliography{lightning2}

\begin{thebibliography}{55}
\providecommand{\natexlab}[1]{#1}
\providecommand{\url}[1]{\texttt{#1}}
\expandafter\ifx\csname urlstyle\endcsname\relax
  \providecommand{\doi}[1]{doi: #1}\else
  \providecommand{\doi}{doi: \begingroup \urlstyle{rm}\Url}\fi

\bibitem[Biderman et~al.(2023)Biderman, Schoelkopf, Anthony, Bradley, O'Brien, Hallahan, Khan, Purohit, Prashanth, Raff, Skowron, Sutawika, and van~der Wal]{biderman2023pythia}
Biderman, S., Schoelkopf, H., Anthony, Q., Bradley, H., O'Brien, K., Hallahan, E., Khan, M.~A., Purohit, S., Prashanth, U.~S., Raff, E., Skowron, A., Sutawika, L., and van~der Wal, O.
\newblock Pythia: A suite for analyzing large language models across training and scaling, 2023.

\bibitem[Bisk et~al.(2019)Bisk, Zellers, Bras, Gao, and Choi]{bisk2019piqa}
Bisk, Y., Zellers, R., Bras, R.~L., Gao, J., and Choi, Y.
\newblock Piqa: Reasoning about physical commonsense in natural language, 2019.

\bibitem[Brown et~al.(2020)Brown, Mann, Ryder, Subbiah, Kaplan, Dhariwal, Neelakantan, Shyam, Sastry, Askell, et~al.]{brown2020language}
Brown, T., Mann, B., Ryder, N., Subbiah, M., Kaplan, J.~D., Dhariwal, P., Neelakantan, A., Shyam, P., Sastry, G., Askell, A., et~al.
\newblock Language models are few-shot learners.
\newblock \emph{Advances in neural information processing systems}, 33:\penalty0 1877--1901, 2020.

\bibitem[Chen et~al.(2023)Chen, Wong, Chen, and Tian]{chen2023extending}
Chen, S., Wong, S., Chen, L., and Tian, Y.
\newblock Extending context window of large language models via positional interpolation, 2023.

\bibitem[Chi et~al.(2022)Chi, Fan, Ramadge, and Rudnicky]{chi2022kerple}
Chi, T.-C., Fan, T.-H., Ramadge, P.~J., and Rudnicky, A.~I.
\newblock Kerple: Kernelized relative positional embedding for length extrapolation, 2022.

\bibitem[Chi et~al.(2023)Chi, Fan, Rudnicky, and Ramadge]{chi2023dissecting}
Chi, T.-C., Fan, T.-H., Rudnicky, A.~I., and Ramadge, P.~J.
\newblock Dissecting transformer length extrapolation via the lens of receptive field analysis, 2023.

\bibitem[Choromanski et~al.(2020)Choromanski, Likhosherstov, Dohan, Song, Gane, Sarl{\'o}s, Hawkins, Davis, Mohiuddin, Kaiser, Belanger, Colwell, and Weller]{Choromanski2020RethinkingAW}
Choromanski, K., Likhosherstov, V., Dohan, D., Song, X., Gane, A., Sarl{\'o}s, T., Hawkins, P., Davis, J., Mohiuddin, A., Kaiser, L., Belanger, D., Colwell, L.~J., and Weller, A.
\newblock Rethinking attention with performers.
\newblock \emph{ArXiv}, abs/2009.14794, 2020.

\bibitem[Clark et~al.(2019)Clark, Lee, Chang, Kwiatkowski, Collins, and Toutanova]{clark2019boolq}
Clark, C., Lee, K., Chang, M.-W., Kwiatkowski, T., Collins, M., and Toutanova, K.
\newblock Boolq: Exploring the surprising difficulty of natural yes/no questions, 2019.

\bibitem[Clark et~al.(2018)Clark, Cowhey, Etzioni, Khot, Sabharwal, Schoenick, and Tafjord]{clark2018think}
Clark, P., Cowhey, I., Etzioni, O., Khot, T., Sabharwal, A., Schoenick, C., and Tafjord, O.
\newblock Think you have solved question answering? try arc, the ai2 reasoning challenge, 2018.

\bibitem[Dao(2023)]{dao2023flashattention2}
Dao, T.
\newblock Flashattention-2: Faster attention with better parallelism and work partitioning.
\newblock \emph{arXiv preprint arXiv:2307.08691}, 2023.

\bibitem[Dao et~al.(2022)Dao, Fu, Ermon, Rudra, and R{\'e}]{dao2022flashattention}
Dao, T., Fu, D.~Y., Ermon, S., Rudra, A., and R{\'e}, C.
\newblock Flash{A}ttention: Fast and memory-efficient exact attention with {IO}-awareness.
\newblock In \emph{Advances in Neural Information Processing Systems}, 2022.

\bibitem[Gao et~al.(2023)Gao, Tow, Abbasi, Biderman, Black, DiPofi, Foster, Golding, Hsu, Le~Noac'h, Li, McDonell, Muennighoff, Ociepa, Phang, Reynolds, Schoelkopf, Skowron, Sutawika, Tang, Thite, Wang, Wang, and Zou]{eval-harness}
Gao, L., Tow, J., Abbasi, B., Biderman, S., Black, S., DiPofi, A., Foster, C., Golding, L., Hsu, J., Le~Noac'h, A., Li, H., McDonell, K., Muennighoff, N., Ociepa, C., Phang, J., Reynolds, L., Schoelkopf, H., Skowron, A., Sutawika, L., Tang, E., Thite, A., Wang, B., Wang, K., and Zou, A.
\newblock A framework for few-shot language model evaluation, 12 2023.
\newblock URL \url{https://zenodo.org/records/10256836}.

\bibitem[Hao et~al.(2024)Hao, Mao, He, Han, Dai, and Zhong]{hao2023improving}
Hao, D., Mao, Y., He, B., Han, X., Dai, Y., and Zhong, Y.
\newblock Improving audio-visual segmentation with bidirectional generation.
\newblock In \emph{Proceedings of the AAAI Conference on Artificial Intelligence}, 2024.

\bibitem[Hendrycks et~al.(2021)Hendrycks, Burns, Basart, Zou, Mazeika, Song, and Steinhardt]{hendrycks2021measuring}
Hendrycks, D., Burns, C., Basart, S., Zou, A., Mazeika, M., Song, D., and Steinhardt, J.
\newblock Measuring massive multitask language understanding, 2021.

\bibitem[Hua et~al.(2022)Hua, Dai, Liu, and Le]{hua2022transformer}
Hua, W., Dai, Z., Liu, H., and Le, Q.~V.
\newblock Transformer quality in linear time.
\newblock \emph{arXiv preprint arXiv:2202.10447}, 2022.

\bibitem[Huang et~al.(2023)Huang, Bai, Zhu, Zhang, Zhang, Su, Liu, Lv, Zhang, Lei, Fu, Sun, and He]{huang2023ceval}
Huang, Y., Bai, Y., Zhu, Z., Zhang, J., Zhang, J., Su, T., Liu, J., Lv, C., Zhang, Y., Lei, J., Fu, Y., Sun, M., and He, J.
\newblock C-eval: A multi-level multi-discipline chinese evaluation suite for foundation models, 2023.

\bibitem[Katharopoulos et~al.(2020{\natexlab{a}})Katharopoulos, Vyas, Pappas, and Fleuret]{katharopoulos2020transformers}
Katharopoulos, A., Vyas, A., Pappas, N., and Fleuret, F.
\newblock Transformers are rnns: Fast autoregressive transformers with linear attention.
\newblock In \emph{International Conference on Machine Learning}, pp.\  5156--5165. PMLR, 2020{\natexlab{a}}.

\bibitem[Katharopoulos et~al.(2020{\natexlab{b}})Katharopoulos, Vyas, Pappas, and Fleuret]{xfmrsarernns}
Katharopoulos, A., Vyas, A., Pappas, N., and Fleuret, F.
\newblock Transformers are rnns: Fast autoregressive transformers with linear attention.
\newblock In \emph{Proceedings of the 37th International Conference on Machine Learning, {ICML} 2020, 13-18 July 2020, Virtual Event}, volume 119 of \emph{Proceedings of Machine Learning Research}, pp.\  5156--5165. {PMLR}, 2020{\natexlab{b}}.
\newblock URL \url{http://proceedings.mlr.press/v119/katharopoulos20a.html}.

\bibitem[Ke et~al.(2021)Ke, He, and Liu]{ke2021rethinking}
Ke, G., He, D., and Liu, T.-Y.
\newblock Rethinking positional encoding in language pre-training.
\newblock In \emph{International Conference on Learning Representations}, 2021.
\newblock URL \url{https://openreview.net/forum?id=09-528y2Fgf}.

\bibitem[Li et~al.(2022)Li, Li, Xiong, and Hoi]{pmlr-v162-li22n}
Li, J., Li, D., Xiong, C., and Hoi, S.
\newblock {BLIP}: Bootstrapping language-image pre-training for unified vision-language understanding and generation.
\newblock In Chaudhuri, K., Jegelka, S., Song, L., Szepesvari, C., Niu, G., and Sabato, S. (eds.), \emph{Proceedings of the 39th International Conference on Machine Learning}, volume 162 of \emph{Proceedings of Machine Learning Research}, pp.\  12888--12900. PMLR, 17--23 Jul 2022.
\newblock URL \url{https://proceedings.mlr.press/v162/li22n.html}.

\bibitem[Li et~al.(2023{\natexlab{a}})Li, Li, Savarese, and Hoi]{li2023blip2}
Li, J., Li, D., Savarese, S., and Hoi, S.
\newblock Blip-2: Bootstrapping language-image pre-training with frozen image encoders and large language models.
\newblock In \emph{arXiv}, 2023{\natexlab{a}}.

\bibitem[Li et~al.(2023{\natexlab{b}})Li, Li, Li, Wang, Jie, and Zhong]{li-etal-2023-map}
Li, W., Li, D., Li, W., Wang, Y., Jie, H., and Zhong, Y.
\newblock {MAP}: Low-data regime multimodal learning with adapter-based pre-training and prompting.
\newblock In Breitholtz, E., Lappin, S., Loaiciga, S., Ilinykh, N., and Dobnik, S. (eds.), \emph{Proceedings of the 2023 CLASP Conference on Learning with Small Data (LSD)}, pp.\  185--190, Gothenburg, Sweden, September 2023{\natexlab{b}}. Association for Computational Linguistics.
\newblock URL \url{https://aclanthology.org/2023.clasp-1.19}.

\bibitem[Liu et~al.(2023)Liu, Li, Wu, and Lee]{liu2023visual}
Liu, H., Li, C., Wu, Q., and Lee, Y.~J.
\newblock Visual instruction tuning.
\newblock In \emph{arXiv}, 2023.

\bibitem[Lu et~al.(2022)Lu, Liu, Wang, Sun, Qin, Li, Shen, Deng, Han, Dai, and Zhong]{lu2022linear}
Lu, K., Liu, Z., Wang, J., Sun, W., Qin, Z., Li, D., Shen, X., Deng, H., Han, X., Dai, Y., and Zhong, Y.
\newblock Linear video transformer with feature fixation, 2022.

\bibitem[Mao et~al.(2023)Mao, Zhang, Xiang, Zhong, and Dai]{Mao_2023_ICCV}
Mao, Y., Zhang, J., Xiang, M., Zhong, Y., and Dai, Y.
\newblock Multimodal variational auto-encoder based audio-visual segmentation.
\newblock In \emph{Proceedings of the IEEE/CVF International Conference on Computer Vision (ICCV)}, pp.\  954--965, October 2023.

\bibitem[Mihaylov et~al.(2018)Mihaylov, Clark, Khot, and Sabharwal]{mihaylov2018suit}
Mihaylov, T., Clark, P., Khot, T., and Sabharwal, A.
\newblock Can a suit of armor conduct electricity? a new dataset for open book question answering, 2018.

\bibitem[Paszke et~al.(2019)Paszke, Gross, Massa, Lerer, Bradbury, Chanan, Killeen, Lin, Gimelshein, Antiga, et~al.]{paszke2019pytorch}
Paszke, A., Gross, S., Massa, F., Lerer, A., Bradbury, J., Chanan, G., Killeen, T., Lin, Z., Gimelshein, N., Antiga, L., et~al.
\newblock Pytorch: An imperative style, high-performance deep learning library.
\newblock \emph{Advances in neural information processing systems}, 32, 2019.

\bibitem[Peng et~al.(2023)Peng, Alcaide, Anthony, Albalak, Arcadinho, Biderman, Cao, Cheng, Chung, Derczynski, Du, Grella, Gv, He, Hou, Kazienko, Kocon, Kong, Koptyra, Lau, Lin, Mantri, Mom, Saito, Song, Tang, Wind, Wo{\'z}niak, Zhang, Zhou, Zhu, and Zhu]{peng-etal-2023-rwkv}
Peng, B., Alcaide, E., Anthony, Q., Albalak, A., Arcadinho, S., Biderman, S., Cao, H., Cheng, X., Chung, M., Derczynski, L., Du, X., Grella, M., Gv, K., He, X., Hou, H., Kazienko, P., Kocon, J., Kong, J., Koptyra, B., Lau, H., Lin, J., Mantri, K. S.~I., Mom, F., Saito, A., Song, G., Tang, X., Wind, J., Wo{\'z}niak, S., Zhang, Z., Zhou, Q., Zhu, J., and Zhu, R.-J.
\newblock {RWKV}: Reinventing {RNN}s for the transformer era.
\newblock In Bouamor, H., Pino, J., and Bali, K. (eds.), \emph{Findings of the Association for Computational Linguistics: EMNLP 2023}, pp.\  14048--14077, Singapore, December 2023. Association for Computational Linguistics.
\newblock \doi{10.18653/v1/2023.findings-emnlp.936}.
\newblock URL \url{https://aclanthology.org/2023.findings-emnlp.936}.

\bibitem[Peng et~al.(2021)Peng, Pappas, Yogatama, Schwartz, Smith, and Kong]{rfa}
Peng, H., Pappas, N., Yogatama, D., Schwartz, R., Smith, N.~A., and Kong, L.
\newblock Random feature attention.
\newblock In \emph{9th International Conference on Learning Representations, {ICLR} 2021, Virtual Event, Austria, May 3-7, 2021}. OpenReview.net, 2021.
\newblock URL \url{https://openreview.net/forum?id=QtTKTdVrFBB}.

\bibitem[Press et~al.(2022)Press, Smith, and Lewis]{alibi}
Press, O., Smith, N., and Lewis, M.
\newblock Train short, test long: Attention with linear biases enables input length extrapolation.
\newblock In \emph{International Conference on Learning Representations}, 2022.
\newblock URL \url{https://openreview.net/forum?id=R8sQPpGCv0}.

\bibitem[Qin et~al.(2022{\natexlab{a}})Qin, Han, Sun, Li, Kong, Barnes, and Zhong]{qin-etal-2022-devil}
Qin, Z., Han, X., Sun, W., Li, D., Kong, L., Barnes, N., and Zhong, Y.
\newblock The devil in linear transformer.
\newblock In \emph{Proceedings of the 2022 Conference on Empirical Methods in Natural Language Processing}, pp.\  7025--7041, Abu Dhabi, United Arab Emirates, December 2022{\natexlab{a}}. Association for Computational Linguistics.
\newblock URL \url{https://aclanthology.org/2022.emnlp-main.473}.

\bibitem[Qin et~al.(2022{\natexlab{b}})Qin, Sun, Deng, Li, Wei, Lv, Yan, Kong, and Zhong]{zhen2022cosformer}
Qin, Z., Sun, W., Deng, H., Li, D., Wei, Y., Lv, B., Yan, J., Kong, L., and Zhong, Y.
\newblock cosformer: Rethinking softmax in attention.
\newblock In \emph{International Conference on Learning Representations}, 2022{\natexlab{b}}.
\newblock URL \url{https://openreview.net/forum?id=Bl8CQrx2Up4}.

\bibitem[Qin et~al.(2023{\natexlab{a}})Qin, Han, Sun, He, Li, Li, Dai, Kong, and Zhong]{qin2023toeplitz}
Qin, Z., Han, X., Sun, W., He, B., Li, D., Li, D., Dai, Y., Kong, L., and Zhong, Y.
\newblock Toeplitz neural network for sequence modeling.
\newblock In \emph{The Eleventh International Conference on Learning Representations}, 2023{\natexlab{a}}.
\newblock URL \url{https://openreview.net/forum?id=IxmWsm4xrua}.

\bibitem[Qin et~al.(2023{\natexlab{b}})Qin, Li, Sun, Sun, Shen, Han, Wei, Lv, Yuan, Luo, et~al.]{qin2023scaling}
Qin, Z., Li, D., Sun, W., Sun, W., Shen, X., Han, X., Wei, Y., Lv, B., Yuan, F., Luo, X., et~al.
\newblock Scaling transnormer to 175 billion parameters.
\newblock \emph{arXiv preprint arXiv:2307.14995}, 2023{\natexlab{b}}.

\bibitem[Qin et~al.(2023{\natexlab{c}})Qin, Sun, Lu, Deng, Li, Han, Dai, Kong, and Zhong]{qin2023linearized}
Qin, Z., Sun, W., Lu, K., Deng, H., Li, D., Han, X., Dai, Y., Kong, L., and Zhong, Y.
\newblock Linearized relative positional encoding.
\newblock \emph{Transactions on Machine Learning Research}, 2023{\natexlab{c}}.

\bibitem[Qin et~al.(2023{\natexlab{d}})Qin, Yang, and Zhong]{qin2023hierarchically}
Qin, Z., Yang, S., and Zhong, Y.
\newblock Hierarchically gated recurrent neural network for sequence modeling.
\newblock In \emph{NeurIPS}, 2023{\natexlab{d}}.

\bibitem[Qin et~al.(2024)Qin, Zhong, and Deng]{qin2023exploring}
Qin, Z., Zhong, Y., and Deng, H.
\newblock Exploring transformer extrapolation.
\newblock In \emph{Proceedings of the AAAI Conference on Artificial Intelligence}, 2024.

\bibitem[Radford et~al.(2021)Radford, Kim, Hallacy, Ramesh, Goh, Agarwal, Sastry, Askell, Mishkin, Clark, Krueger, and Sutskever]{radford2021learning}
Radford, A., Kim, J.~W., Hallacy, C., Ramesh, A., Goh, G., Agarwal, S., Sastry, G., Askell, A., Mishkin, P., Clark, J., Krueger, G., and Sutskever, I.
\newblock Learning transferable visual models from natural language supervision.
\newblock In \emph{arXiv}, 2021.

\bibitem[Sakaguchi et~al.(2019)Sakaguchi, Bras, Bhagavatula, and Choi]{sakaguchi2019winogrande}
Sakaguchi, K., Bras, R.~L., Bhagavatula, C., and Choi, Y.
\newblock Winogrande: An adversarial winograd schema challenge at scale, 2019.

\bibitem[Sap et~al.(2019)Sap, Rashkin, Chen, LeBras, and Choi]{sap2019socialiqa}
Sap, M., Rashkin, H., Chen, D., LeBras, R., and Choi, Y.
\newblock Socialiqa: Commonsense reasoning about social interactions, 2019.

\bibitem[Shen et~al.(2023)Shen, Li, Zhou, Qin, He, Han, Li, Dai, Kong, Wang, Qiao, and Zhong]{Shen_2023_CVPR}
Shen, X., Li, D., Zhou, J., Qin, Z., He, B., Han, X., Li, A., Dai, Y., Kong, L., Wang, M., Qiao, Y., and Zhong, Y.
\newblock Fine-grained audible video description.
\newblock In \emph{Proceedings of the IEEE/CVF Conference on Computer Vision and Pattern Recognition (CVPR)}, pp.\  10585--10596, June 2023.

\bibitem[Su et~al.(2021)Su, Lu, Pan, Wen, and Liu]{su2021roformer}
Su, J., Lu, Y., Pan, S., Wen, B., and Liu, Y.
\newblock Roformer: Enhanced transformer with rotary position embedding.
\newblock \emph{arXiv preprint arXiv:2104.09864}, 2021.

\bibitem[Sun et~al.(2023{\natexlab{a}})Sun, Qin, Deng, Wang, Zhang, Zhang, Barnes, Birchfield, Kong, and Zhong]{10149455}
Sun, W., Qin, Z., Deng, H., Wang, J., Zhang, Y., Zhang, K., Barnes, N., Birchfield, S., Kong, L., and Zhong, Y.
\newblock Vicinity vision transformer.
\newblock \emph{IEEE Transactions on Pattern Analysis and Machine Intelligence}, 45\penalty0 (10):\penalty0 12635--12649, 2023{\natexlab{a}}.
\newblock \doi{10.1109/TPAMI.2023.3285569}.

\bibitem[Sun et~al.(2023{\natexlab{b}})Sun, Dong, Huang, Ma, Xia, Xue, Wang, and Wei]{2307.08621}
Sun, Y., Dong, L., Huang, S., Ma, S., Xia, Y., Xue, J., Wang, J., and Wei, F.
\newblock Retentive network: A successor to transformer for large language models, 2023{\natexlab{b}}.

\bibitem[Tillet et~al.(2019)Tillet, Kung, and Cox]{Tillet2019TritonAI}
Tillet, P., Kung, H.-T., and Cox, D.~D.
\newblock Triton: an intermediate language and compiler for tiled neural network computations.
\newblock \emph{Proceedings of the 3rd ACM SIGPLAN International Workshop on Machine Learning and Programming Languages}, 2019.

\bibitem[Touvron et~al.(2023{\natexlab{a}})Touvron, Lavril, Izacard, Martinet, Lachaux, Lacroix, Rozi{\`e}re, Goyal, Hambro, Azhar, Rodriguez, Joulin, Grave, and Lample]{touvron2023llama}
Touvron, H., Lavril, T., Izacard, G., Martinet, X., Lachaux, M.-A., Lacroix, T., Rozi{\`e}re, B., Goyal, N., Hambro, E., Azhar, F., Rodriguez, A., Joulin, A., Grave, E., and Lample, G.
\newblock Llama: Open and efficient foundation language models.
\newblock \emph{arXiv preprint arXiv:2302.13971}, 2023{\natexlab{a}}.

\bibitem[Touvron et~al.(2023{\natexlab{b}})Touvron, Martin, Stone, Albert, Almahairi, Babaei, Bashlykov, Batra, Bhargava, Bhosale, Bikel, Blecher, Ferrer, Chen, Cucurull, Esiobu, Fernandes, Fu, Fu, Fuller, Gao, Goswami, Goyal, Hartshorn, Hosseini, Hou, Inan, Kardas, Kerkez, Khabsa, Kloumann, Korenev, Koura, Lachaux, Lavril, Lee, Liskovich, Lu, Mao, Martinet, Mihaylov, Mishra, Molybog, Nie, Poulton, Reizenstein, Rungta, Saladi, Schelten, Silva, Smith, Subramanian, Tan, Tang, Taylor, Williams, Kuan, Xu, Yan, Zarov, Zhang, Fan, Kambadur, Narang, Rodriguez, Stojnic, Edunov, and Scialom]{2307.09288}
Touvron, H., Martin, L., Stone, K., Albert, P., Almahairi, A., Babaei, Y., Bashlykov, N., Batra, S., Bhargava, P., Bhosale, S., Bikel, D., Blecher, L., Ferrer, C.~C., Chen, M., Cucurull, G., Esiobu, D., Fernandes, J., Fu, J., Fu, W., Fuller, B., Gao, C., Goswami, V., Goyal, N., Hartshorn, A., Hosseini, S., Hou, R., Inan, H., Kardas, M., Kerkez, V., Khabsa, M., Kloumann, I., Korenev, A., Koura, P.~S., Lachaux, M.-A., Lavril, T., Lee, J., Liskovich, D., Lu, Y., Mao, Y., Martinet, X., Mihaylov, T., Mishra, P., Molybog, I., Nie, Y., Poulton, A., Reizenstein, J., Rungta, R., Saladi, K., Schelten, A., Silva, R., Smith, E.~M., Subramanian, R., Tan, X.~E., Tang, B., Taylor, R., Williams, A., Kuan, J.~X., Xu, P., Yan, Z., Zarov, I., Zhang, Y., Fan, A., Kambadur, M., Narang, S., Rodriguez, A., Stojnic, R., Edunov, S., and Scialom, T.
\newblock Llama 2: Open foundation and fine-tuned chat models, 2023{\natexlab{b}}.

\bibitem[Vaswani et~al.(2017)Vaswani, Shazeer, Parmar, Uszkoreit, Jones, Gomez, Kaiser, and Polosukhin]{vaswani2017attention}
Vaswani, A., Shazeer, N., Parmar, N., Uszkoreit, J., Jones, L., Gomez, A.~N., Kaiser, {\L}., and Polosukhin, I.
\newblock Attention is all you need.
\newblock \emph{Advances in neural information processing systems}, 30, 2017.

\bibitem[Xiao et~al.(2023)Xiao, Tian, Chen, Han, and Lewis]{xiao2023efficient}
Xiao, G., Tian, Y., Chen, B., Han, S., and Lewis, M.
\newblock Efficient streaming language models with attention sinks, 2023.

\bibitem[Yang et~al.(2023)Yang, Wang, Shen, Panda, and Kim]{2312.06635}
Yang, S., Wang, B., Shen, Y., Panda, R., and Kim, Y.
\newblock Gated linear attention transformers with hardware-efficient training, 2023.

\bibitem[Zellers et~al.(2019)Zellers, Holtzman, Bisk, Farhadi, and Choi]{zellers2019hellaswag}
Zellers, R., Holtzman, A., Bisk, Y., Farhadi, A., and Choi, Y.
\newblock Hellaswag: Can a machine really finish your sentence?, 2019.

\bibitem[Zhang et~al.(2022)Zhang, Roller, Goyal, Artetxe, Chen, Chen, Dewan, Diab, Li, Lin, Mihaylov, Ott, Shleifer, Shuster, Simig, Koura, Sridhar, Wang, and Zettlemoyer]{zhang2022opt}
Zhang, S., Roller, S., Goyal, N., Artetxe, M., Chen, M., Chen, S., Dewan, C., Diab, M., Li, X., Lin, X.~V., Mihaylov, T., Ott, M., Shleifer, S., Shuster, K., Simig, D., Koura, P.~S., Sridhar, A., Wang, T., and Zettlemoyer, L.
\newblock Opt: Open pre-trained transformer language models, 2022.

\bibitem[Zheng et~al.(2022)Zheng, Wang, and Kong]{zheng2022linear}
Zheng, L., Wang, C., and Kong, L.
\newblock Linear complexity randomized self-attention mechanism.
\newblock In \emph{International Conference on Machine Learning}, pp.\  27011--27041. PMLR, 2022.

\bibitem[Zheng et~al.(2023)Zheng, Yuan, Wang, and Kong]{zheng2023efficient}
Zheng, L., Yuan, J., Wang, C., and Kong, L.
\newblock Efficient attention via control variates.
\newblock In \emph{International Conference on Learning Representations}, 2023.
\newblock URL \url{https://openreview.net/forum?id=G-uNfHKrj46}.

\bibitem[Zhou et~al.(2023)Zhou, Shen, Wang, Zhang, Sun, Zhang, Birchfield, Guo, Kong, Wang, and Zhong]{zhou2023audiovisual}
Zhou, J., Shen, X., Wang, J., Zhang, J., Sun, W., Zhang, J., Birchfield, S., Guo, D., Kong, L., Wang, M., and Zhong, Y.
\newblock Audio-visual segmentation with semantics, 2023.

\end{thebibliography}
\bibliographystyle{icml2024}

\end{document}